\documentclass[letterpaper, 10 pt, conference]{ieeeconf}

\IEEEoverridecommandlockouts    
\usepackage{times}

\usepackage{multicol}
\usepackage[bookmarks=true]{hyperref}

\usepackage{amsmath,amssymb,bbm,bm,mathtools}
\usepackage[ruled,vlined]{algorithm2e}
\usepackage{comment}
\usepackage{url}
\usepackage{cite}
\usepackage{booktabs}
\usepackage{ntheorem}
\usepackage{thmtools}
\usepackage{graphicx} % for pdf, bitmapped graphics files
\usepackage{epsfig} % for postscript graphics files
\usepackage[format=plain,textfont=footnotesize, 
labelfont=footnotesize,
tableposition=top,justification=justified,belowskip=-3pt,aboveskip=8pt]{caption}
\usepackage{xcolor}

\usepackage{commands}
\def\isconfidential{0}  % Set to 1 if confidential, or 0
\newif\ifconfidential

\confidentialtrue
\if\isconfidential1
\usepackage{fancyhdr}
\fi
\newcommand{\makeconfidential}{
\if\isconfidential1
\fancyhead[C]{{\bf MERL CONFIDENTIAL}. $\copyright$MERL, $2024$. }
\fancyfoot[C]{{\bf MERL CONFIDENTIAL}. $\copyright$MERL, $2024$}
\thispagestyle{fancy}
\pagestyle{fancy}
\else
\thispagestyle{empty}
\pagestyle{empty}
\fi
}

\begin{document}

% paper title
\title{\LARGE \bf Chance-Constrained Information-Theoretic Stochastic Model Predictive Control with Safety Shielding}

% You will get a Paper-ID when submitting a pdf file to the conference system
\author{Ji Yin$^1$, Panagiotis Tsiotras$^1$, Karl Berntorp$^\star$
\thanks{$^{1}$Ji Yin and Panagiotis Tsiotras are with D. Guggenheim School of Aerospace Engineering, Georgia Institute of Technology, GA. This work has been partially supported by ONR award N00014-23-1-2353 and NSF award CPS-2211755. Ji Yin  was an intern at MERL at the time of this research and this support was not applied to the work at MERL.
        {\tt\footnotesize \{jyin81,tsiotras\}@gatech.edu}}%
\thanks{$^\star$ Corresponding author: Karl Berntorp, Mitsubishi Electric Research Labs (MERL), 
Cambridge, MA, USA. Email: {\tt \small karl.o.berntorp@ieee.org}.}
}

\maketitle
\makeconfidential
\begin{abstract}
This paper introduces a novel nonlinear stochastic model predictive control path integral (MPPI)  method, which considers chance constraints on system states. The proposed belief-space stochastic MPPI (BSS-MPPI) applies Monte-Carlo sampling to evaluate state distributions resulting from underlying systematic disturbances, and utilizes a Control Barrier Function (CBF) inspired heuristic in belief space to fulfill the specified chance constraints. 
Compared to several previous stochastic predictive control methods, our approach applies to general nonlinear dynamics without requiring the computationally expensive system linearization step. Moreover, the BSS-MPPI controller can solve optimization problems without limiting the form of the objective function and chance constraints. 
By multi-threading the sampling process using a GPU, we can achieve fast real-time planning for time- and safety-critical tasks such as autonomous racing. Our results on a realistic race-car simulation study show significant reductions in  constraint violation compared to some of the prior MPPI approaches, while being comparable in computation times.
\end{abstract}

%\IEEEpeerreviewmaketitle

\section{Introduction}\label{sec:Introduction}

Safety is a pivotal concern in the expanding global robotics industry, directly influencing the range of robotic applications. Hence, the importance of safety in robot applications cannot be understated and it is paramount when humans interact with robots, which operate autonomously, often with unpredictable motions. This unpredictability necessitates stringent safety measures to prevent any potential harm to individuals, especially those unaccustomed to interacting with robots \cite{tzafestas2014introduction}.

Over a span of 30 years, 37 robot-related accidents were reported, with 27 incidents resulting in a worker's death between 1984 and 2013. This data underscore the necessity of safety protocols to prevent workplace fatalities~\cite{RobotSafety}. 
The evolution of robotic systems from performing static manipulation tasks into more collaborative and dynamic,  further emphasizes the need for robust safety processes~\cite{RobotSafetyProcess}. 
To address safety and increase robot reliability, it is crucial to consider uncertainties in robot planning and control. Uncertainties, such as dynamical disturbances, can lead to unpredictable robot behavior, which, in turn, might pose risks to human operators or other robots working in proximity. For instance, an autonomous vehicle might fail to respond adequately to unexpected road conditions, endangering the safety of its occupants and other road users.

Many control systems are designed based on deterministic models of the real robot dynamics; these controllers may fail to function as intended due to unexpected disturbances, leading to sub-optimal or even unsafe operations. To this end, during the last decade numerous efforts have been made towards control design incorporating model and environment uncertainties, for example, in a stochastic model-predictive control (SMPC) framework \cite{SMPCSurvey}. However, the ability to handle nonlinear dynamics and/or non-Gaussian uncertainties in a computationally efficient manner is still a concern.

We propose a novel control approach that considers chance constraints in belief space, using path integral and control barrier functions (CBFs) theories. 
Specifically, we design a heuristic in belief space inspired by \textit{discrete-time} CBFs (DCBFs) to push the probability of collision lower than a user-specified threshold and integrate it with the Shield Model Predictive Path Integral (Shield-MPPI) of \cite{ShieldMPPI} to achieve real-time planning. 
We call the resulting control method the belief-space stochastic MPPI (BSS-MPPI). By using Monte-Carlo sampling to evaluate the state distributions, we estimate a general distribution and extract the  first two moments into the state-dependent part of the objective function of the optimization problem, and solve for a solution fulfilling chance constraints for safety. However, BSS-MPPI can also integrate with other, nonsampling based uncertainty-propagation schemes, such as nonlinear Kalman filters. 

Compared to most previous trajectory optimization methods that consider chance constraints, our approach differs in the following ways:
\begin{enumerate}
    \item[a)] 
    First, BSS-MPPI solves a nonlinear stochastic optimal control problem by forward simulation, hence avoiding the computationally-expensive explicit optimization and linearization steps. It can achieve real-time planning on either CPU or GPU by multi-threading. 
    \item[b)] 
    Second, BSS-MPPI can be applied to more general nonlinear stochastic systems. Previous works that consider chance constraints mostly assume linear/linearized dynamics, linear constraints, and some specific form of disturbances (e.g., Gaussian), which limit the range of applicability in robotic systems.
    \item[c)] 
    Third, finding a verified nonsaturating CBF (with bounded control input) is difficult for high-dimensional systems \cite{so2023train}. We design CBF-inspired heuristics that provide better scalability at the cost of less formal guarantees and apply them to high-dimensional belief space. 
    % uses control barrier functions to ensure that the actual risk is below the assigned threshold with theoretical guarantee, enabling robust real-time planning.
\end{enumerate}

\subsection{Related Work}

Controlling the evolution of state uncertainties to fulfill safety requirements is not a trivial task. 
There are many works in the literature that study motion planning under uncertainties using chance constraints. 
For instance, distributional (e.g., covariance) steering is a recent technique that is used to steer an initial distribution to a target distribution of target mean and covariance values, while subject to chance constraints. 
For unmanned ground vehicles (UGVs)
In \cite{CSVehiclePathPlanning, CCMPPI}, the authors formulate covariance steering as a convex optimization problem and use its solution to plan trajectories for linearized vehicle models with additive Gaussian noise in challenging environments. For unmanned aerial vehicles (UAVs), \cite{CCOCP_TRO}  utilize covariance control to achieve path planning with obstacles, and \cite{lew2019chance} uses this technique for altitude control of a rocket.
%
% \panos{The lit review is based heavily on our prior work. There is a ton of work on CC planning for robotic systems. If we do not cite those, it gives plenty ammunition to the reviewers to attack the paper}
%
Other related approaches include \cite{ChanceConstrainedSequentialConvexProgramming}, which reformulates obstacle avoidance chance
constraints using the signed distance function and applies the sequential convex programming (SCP) algorithm, and \cite{ChanceConstrainedDP}, which considers the chance-constrained trajectory optimization problem by integrating the constraints into the cost functions and uses dynamic programming to obtain a solution. These approaches all require linearization of the dynamics, or using a discretized state space, which makes real-time planning and replanning computationally challenging, especially for high-dimensional systems.
The authors of \cite{CLARK2021109688} and \cite{so2023almostsure} propose the use of stochastic CBFs (SCBF), which provide $100\%$ safety assurance for a system with Brownian noise in continuous space. 
However, the theoretical guarantees of their approach include an implicit assumption that the control inputs and the dynamical system must be unbounded, which is not practical for many robotic systems.

Compared to many other nonlinear controllers, such as SCP, iLQR, and iLQG, MPPI implements the optimal control problem by forward simulation using the original nonlinear dynamics  and hence avoids the linearizations  involved in the explicit optimal control solvers. MPPI does not restrict the form of the objective function, which can be nonconvex and discrete. However, the base MPPI uses only deterministic dynamics. To improve the robustness of MPPI, many variants of the algorithm have been developed \cite{lee2023learning, kim2024adaptive, hatch2021value}.   In particular, \cite{RAMPPI, UncertaintyPushingMPPI} introduce penalty costs for constraint violations due to dynamical and environmental uncertainties to the MPPI objective function, but they do not provide theoretical guarantee for probablistic constraint satisfaction. The work \cite{CSMPPI} also uses chance constraints, but it assumes linear systems and Gaussian disturbances, hence resulting in Gaussian posterior densities, which are restrictive.
%\subsection{Contributions}
% \subsection{Notation}
% \kbcomment{fill in notation here}

\section{Shield-MPPI Review} 
Consider a general nonlinear dynamical system,
\begin{align}
 x_{k+1} = f(x_k, u_k),\label{DeterministicNonlinearSystem}   
\end{align}
where $x_k \in \mathcal{D} \subseteq \Re^{n_x}$ is the state, $u_k \in \Re^{n_u} \sim \mathcal{N}(v_k, \Sigma_{\epsilon})$ is the control input with the mean control $v_k$ at time step $k=0,\hdots,K-1$ and constant covariance $\Sigma_{\epsilon}$. Let $\phi(x)$ denote the terminal state cost, $q(x)$ the step state cost, and $\lambda$ the weight for control cost. Then, the Shield MPPI (S-MPPI) controller solves the following problem,
\begin{align}\label{eqn:Shield MPPI objective}
    \min_\vv J(\vv) &= \nonumber \\
    &\mathbb{E}\left[ \phi(x_K) + \sum^{K-1}_{k=0} \left(q(x_k)  + \frac{\lambda}{2}v_k^\intercal \Sigma^{-1}_\epsilon v_k \right) \right],
\end{align}
subject to \eqref{DeterministicNonlinearSystem} with initial condition $x_0 = x(0)$, and the first-order DCBF safety condition,
\begin{align}\label{eqn:DCBF_constraint1}
     \sup_{v \in \mathcal{U}}h(f(x_k, v_k)) - h(x_k) \geq - p (h(x_k)),
\end{align}
for $k = 0, \hdots, K-1$, where the continuous function $h(\cdot): \mathbb{R}^n \rightarrow \mathbb{R}$ is the DCBF that defines a safe set in the state space,
\begin{align}\label{eqn:safeset}
    \mathcal{S}\coloneqq\{x \in \mathcal{D}| h(x) \geq 0\}.
\end{align}
The function $p(\cdot)$ is a class-$\kappa$ function that is strictly increasing such that $p(0) = 0$. 
Fulfilling the DCBF constraint \eqref{eqn:DCBF_constraint1} will ensure that the system stays within the safe set $\mathcal{S}$, and the S-MPPI achieves this by minimizing the violation of \eqref{eqn:DCBF_constraint1} using path integral and gradient-based optimizations \cite{ShieldMPPI}.

Compared to the original MPPI \cite{KLMPPI}, S-MPPI has three major advantages. 
First, the vanilla MPPI requires a significant amount of sample trajectories (normally thousands of samples) at each time step for its success, thus 
often requiring the multi-threading computational ability offered by modern GPUs, which are expensive and generally not available on smaller-size robots. 
On the other hand, S-MPPI typically requires an order of magnitude less trajectories to achieve equivalently satisfactory, and even safer performance. 
Second, S-MPPI is less sensitive to cost parameter variations, saving efforts for cost engineering and allowing for more general, less task-specific cost designs. 
Third, S-MPPI is  safer than the MPPI, in general.

Despite all the attractive properties of S-MPPI, it is still a control approach that uses only deterministic dynamics. 
To remedy this, we introduce a CBF-inspired heuristic to S-MPPI, such that the resulting Belief-space MPPI (BSS-MPPI) can consider model uncertainties and plan in the belief space.

\section{Problem Formulation}\label{Section: ProblemFormulation}

In this paper, the objective is to solve the optimal control problem,
\begin{subequations}\label{problem:BSSMPPIProblemformulation}
\begin{align}
&\min_\vv J(\vv) \text{, subject to,} \label{eqn:BSS-MPPIobjective}\\
&x_{k+1} = f(x_k, u_k, w_k),\quad u_k \sim \mathcal{N}(v_t, \Sigma_\epsilon),\label{eqn:DisturbedNonlinearSystem}\\
&\textrm{Pr}(x_k \in \mathcal{F}) > 1 - P_\text{fail}, \text{ for } k = 0, \hdots, K-1, \label{eqn:chance_constraint}\\
&x_0 = x(0),  \label{BSSMPPIInitialCondition}
\end{align}
\end{subequations}
where the dynamical noise $w_k$ in \eqref{eqn:DisturbedNonlinearSystem} can take arbitrary forms,  $\mathcal{F}$ in \eqref{eqn:chance_constraint} is the feasible region, and \eqref{eqn:chance_constraint} denotes the set of chance constraints where the probability of violating the constraint is below a specified threshold value $P_\text{fail}$. In Sec.~\ref{sec:BSS-MPPI} we explain our proposed method, BSS-MPPI, for solving \eqref{problem:BSSMPPIProblemformulation}.

\section{Belief-Space MPPI}\label{sec:BSS-MPPI}

In this section, we briefly review the MPPI control update law and develop the BSS-MPPI algorithm by integrating the chance constraint \eqref{eqn:chance_constraint} using a proposed belief-space heuristic in discrete time.

\subsection{MPPI Control}
The novel BSS-MPPI controller has an objective function in the same form as the original MPPI, thus we can use the MPPI control update law. The MPPI algorithm solves an optimization problem by sampling control inputs and forward simulating a large number of trajectories. Assuming that the algorithm has $M$ trajectory samples with prediction horizon $K$, for the problem \eqref{eqn:Shield MPPI objective} subject to \eqref{DeterministicNonlinearSystem}, the MPPI algorithm has the following control update law,
\begin{equation}\label{eqn:MPPI_update_law}
\vv^+ = \sum^{M}_{m=1}\omega_m \vu^{m}/ \sum^{M}_{m = 1}\omega_m,
\end{equation}
where the $m^{th}$ sample control sequence $\vu^{m} = \{u_0^m, \hdots, u_{K-1}^m \}$, $u^m_t \sim \mathcal{N}(v_k, \Sigma_\epsilon)$,  and where
\begin{equation}\label{eqn:Weights}
    \omega_m = \text{exp}\left(-\frac{1}{\lambda}\left(S_m - \beta \right)\right).
\end{equation}
Note that $\beta = \min_{m=1,\ldots,M} S_m$ in \eqref{eqn:Weights} is introduced to prevent numerical overflow, and $S_m$ is the cost of the $m^{th}$ sample trajectory, given by,
\begin{equation}\label{eqn:TrajectoryCost}
    S_m = \phi(x^{m}_K) + \sum^{K-1}_{k=0}q(x^{m}_k) +
     \gamma (v^m_k)^\intercal \Sigma_\epsilon^{-1} u^{m}_k. 
\end{equation}
\subsection{CBF-inspired heuristic considering chance constraints}
\subsubsection{Safe heuristic in belief space}
To fulfill the chance constraint \eqref{eqn:chance_constraint}, we choose a safe set $\mathcal{S} \subset \mathcal{F}$, such that,
\begin{align}
    \bar{x} \in \mathcal{S} \iff \textrm{Pr}(x \in \mathcal{F}) > 1 - P_\text{fail}, \ P_\text{fail} \in (0,1],
\end{align}
where $\bar{x} = \mathbb{E}[x]$. We introduce the following assumption on the feasible region:
% The disturbed model \eqref{eqn:DisturbedNonlinearSystem} contains white noise $w_k$ with $\mathbb{E}[w_k] = 0$, and we further assume that,
% \begin{align}\label{eqn:Assumption1}
%     f(\bar{x}_k, u_k) = \mathbb{E}[f(x_k, u_k)].
% \end{align}
% It follows from \eqref{eqn:DisturbedNonlinearSystem} and \eqref{eqn:Assumption1} that the disturbance-free model is given by,
% \begin{align}\label{eqn:nominal_dynamics}
%     \bar{x}_{k+1} = f(\bar{x}_k, u_k).
% \end{align}
\begin{ass}\label{ass:1}
    The feasible region $\mathcal{F}$ is defined by the intersection of $Z$ inequalities, such that,
\begin{align}
    \mathcal{F} \coloneqq\ \bigcap^Z_{i=1} \{x: c_i(x) < 0\}, \  i=1,\hdots,Z.
\end{align} 
% \kbcomment{Why do we need this assumption? Why cant we just in (5c) write the constraints as $h_i(x_k) > 0$?? I dont see why we need linear inequalities and can be quite restrictive}
\end{ass}
%
% \kbcomment{I rewrote the following section}

Using Assumption~\ref{ass:1} and  the Boole-Bonferroni inequality~\cite{prekopa1988boole}, any state $x_k$ subject to the constraints,
\begin{subequations}\label{eqn:chance_constraint2a}
\begin{align}
    \text{Pr}(c_i(x) \geq 0) & \leq p_i, \quad i = 1,\hdots, Z,\label{eqn:chance_constraint2}\\
    \sum^Z_{i=1} p_i &\leq P_\text{fail}.
\end{align}
\end{subequations}
also satisfy the original constraint \eqref{eqn:chance_constraint}. From \eqref{eqn:chance_constraint2a}, we can approximate each chance constraint by 
\begin{align} \label{eqn:deterministic_chance_constraint}
    c_i(\bar{x}_k) + \nu_i \sqrt{ \eta_i^\intercal \Sigma_k \eta_i} \leq 0,
\end{align}
where $\Sigma_k \in \mathbb{R}^{n_x \times n_x}$ is the covariance of $x_k$, $\eta_i = \nabla_x c_i(\bar{x}_k)$, and $\nu_i$ is the \emph{back-off coefficient} \cite{Quirynen2021}. We can combine $\bar{x}$ and $\Sigma$ and form a belief-space system,
\begin{align}\label{eqn:BeliefSpaceSystem}
    \hat{z} &= \begin{bmatrix}
        \bar{x}\\
        \text{vec}(\Sigma)
    \end{bmatrix},
\end{align}
where $\hat{z} \in \mathcal{Z} \subseteq \Re^{n_z}$, $n_z = n_x  (n_x+1)$. 
Hence, following from  \eqref{eqn:safeset} and \eqref{eqn:deterministic_chance_constraint}, we obtain the following heuristic,
\begin{align}\label{eqn:DCBFchoice0}
    h_i(\hat{z}) = -c_i(\bar{x}) - \nu_i \sqrt{ \eta_i^\intercal \Sigma \eta_i},
\end{align}
with a superlevel set $\mathcal{S}_i$ containing all states satisfying \eqref{eqn:deterministic_chance_constraint},
\begin{align}\label{eqn:SubSafeSet}
    \mathcal{S}_i = \{\hat{z} \in \mathcal{Z} | h_i(\hat{z}) \geq 0 \}.
\end{align}
Consequently, the overall safe heuristic for Problem \eqref{problem:BSSMPPIProblemformulation} can be chosen as
\begin{align}\label{eqn:DCBFchoice}
    h(\hat{z}) = \min(h_1(\hat{z}),\hdots, h_Z(\hat{z})),
\end{align}
with a corresponding safe set $\mathcal{S} \subseteq \mathcal{Z}$ being the intersection of all superlevel sets $\mathcal{S}_i$ for $i = 1,\hdots,Z$, given by,
\begin{align}\label{eqn:SubSafeSetUnion}
    S = \bigcap^Z_{i=1} S_i .
\end{align}
\begin{rem}\label{rem:1}
The back-off coefficient value $\nu_i$ for each deterministic chance constraint is computed to ensure the probability level $p_i$ in the chance constraint. One option is to use the Cantelli-Chebyshev inequality, that is, $\nu_i = \sqrt{\frac{1-p_i}{p_i}}$, which holds regardless of the underlying probability distribution. However, as a consequence, it may lead to relatively conservative bounds \cite{Telen2015,Quirynen2021,SeanVaskov2024}. Alternatively, for approximately normal-distributed state trajectories, we can set $\nu_i = \sqrt{2}\mathrm{erf}^{-1}(1-2p_i)$ where $\mathrm{erf}^{-1}$ is the inverse error function. The paper \cite{10155999} discusses an alternative, sampling-based approach for constraint tightening.
\end{rem}

\subsubsection{Safety condition}
Assume the belief-space state follows the system $\hat{z}_{k+1} = f_z(\hat{z}_k, v_k)$. The safety condition for the heuristic \eqref{eqn:DCBFchoice} is given by,
\begin{align}\label{eqn:SafeCondition}
    h(\hat{z}_{k+1}) - h(\hat{z}_k) \geq - p (h(\hat{z}_k)),
\end{align}
where $p(\cdot)$ is a class-$\kappa$ function. Satisfying \eqref{eqn:SafeCondition} gives us two properties,
\begin{property}\label{property1}
    Given an initial condition $\hat{z}_0 \in \mathcal{S}$ and a control sequence $\{v_k\}_{k=0}^\infty$ such that all $(\hat{z}_k, v_k)$ pairs satisfy~\eqref{eqn:SafeCondition}, then $\hat{z}_k \in \mathcal{S}$ for all $k \in \mathbb{Z}_{\geq 0}$.
\end{property}
\begin{property}\label{property2}
    Let $\hat{z}_0 \in \mathcal{Z}\setminus \mathcal{S}$ and let
     a control sequence $\{v_k\}_{k=0}^\infty$ such that, 
     for all $k \in \mathbb{Z}_{\geq 0}$, the pair
    $(z_k, v_k)$ satisfies~\eqref{eqn:SafeCondition}.
    Then the
    state $z_k$ converges to the safe set $\mathcal{S}$ asymptotically.
\end{property}
The proofs of both properties are provided in \cite{ShieldMPPI}.
\begin{rem}\label{rem:2}
If there always exists a control $v$ such that \eqref{eqn:SafeCondition} is satisfied for all $\hat{z} \in \mathcal{Z}$, $h(\hat{z})$ becomes a discrete-time control barrier function (DCBF). The HJ reachability analysis, which is commonly used to find verified CBFs, however, is computationally intractable for systems of high dimensions \cite{mitchell2008flexible}. For this reason, finding verified CBFs for belief-space systems is difficult in most cases due to large dimensionality. The proposed BSS-MPPI minimizes violation of \eqref{eqn:SafeCondition} and achieves an average $98.7\%$ satisfaction rate of \eqref{eqn:SafeCondition} in our simulations, effectively keeping system states within the safe set and significantly reduces collision rates.
\end{rem}

% \subsubsection{CBF feasibility challenge}
% As discussed in \cite{DCBFsetDiscussion}, feasibility issues can arise when part of the safe set is not reachable given the current system state and dynamics. 
% It is possible to find a verified CBF for a given system using Hamilton-Jacobi reachability analysis \cite{CBFreachability}, such that feasible controls always exist and the CBF safety condition \eqref{eqn:DCBF_constraint1} holds. 
% However, HJ reachability is computationally intractable for high-dimensional systems (e.g., more than 5 dimensions) \cite{mitchell2008flexible, so2023train}. 
% The proposed,  BSS-MPPI  alleviates the feasibility issue by integrating the CBF safety constraint into the cost evaluation of a simulation-based trajectory sampling process, resulting in control sequences that are more likely to be feasible. Sec.~\ref{sec:safety_aware_objective_function_using_DCBF} discusses more details.

\subsection{Augmented System}

Note that the MPPI objective function \eqref{eqn:Shield MPPI objective} and the corresponding control update law \eqref{eqn:MPPI_update_law}-\eqref{eqn:TrajectoryCost} share the same state-dependent step cost $q(\cdot)$, which can be of  arbitrary form. 
To integrate the DCBF with chance constraints into the MPPI objective function and the control update law, we introduce a system in belief space that includes the mean and covariance of the system \eqref{eqn:DisturbedNonlinearSystem}, then augment the new system to combine two states in one, such that the safety condition \eqref{eqn:SafeCondition} can be included in the step running cost $q(\cdot)$.
\subsubsection{Belief-space system}
We can separate the system state following \eqref{eqn:DisturbedNonlinearSystem} into the mean and disturbed parts,
\begin{align}
    x_{k} = \bar{x}_{k} + \Tilde{x}_{k}.
\end{align}
Given a sequence of sampled controls $\vu = [u_0, \hdots, u_{K-1}]$, the mean state $\bar{x}_k$ follows the nominal system,
\begin{align}
    \bar{x}_{k+1} = \bar{f}(\bar{x}_{k}, u_k) = \mathbb{E}[f(x_k, u_k, w_k)],
\end{align}
and the covariance propagation evolves according to 
\begin{align}
    \Sigma_{x_{k+1}}  = \text{Cov}[f(x_k, u_k, w_k)] = f_\Sigma(\Sigma_{x_k}, u_k).
\end{align}
We can then describe \eqref{eqn:DisturbedNonlinearSystem} using the belief-space system \eqref{eqn:BeliefSpaceSystem},
\begin{subequations}\label{eqn:belief_space_system}
\begin{align}
    \hat{z}_{k+1} &= \begin{bmatrix}
        \bar{x}_{k+1}\\
        \text{vec}(\Sigma_{x_{k+1}})
    \end{bmatrix} 
    = \begin{bmatrix}
        \bar{f}(\bar{x}_k, u_k)\\
        \text{vec}(f_\Sigma(\Sigma_{x_k}, u_k))
    \end{bmatrix}\\
    & = f_z(\hat{z}_k, u_k).
\end{align}
\end{subequations}
In some cases, as demonstraded in our simulations, only part of the uncertainty covariance matrix is needed, thus a smaller state space can be used.
For linear systems with Gaussian additive disturbances and for systems with nonlinearities on a specific form, $\bar{f}$ and $f_\Sigma$ can be expressed analytically \cite{Huber2011}.
For nonlinear systems, data-driven approaches, such as neural nets and Gaussian processes \cite{RLMPPI, CautiousMPC, cao2017gaussian} can be used to model the mean and covariance propagations. In this work, we propose to apply Monte-Carlo sampling to estimate the empirical mean and covariance propagation, such that,
\begin{align}
    \bar{x}_{k+1} = \ExP{}{x_{k+1}} \approx \frac{1}{N}\sum^{N-1}_{n=0}x^n_{k+1},
\end{align}
and,
\begin{align}\label{eqn:Monte-CarloCovarianceEstimation}
    \Sigma_{x_{k+1}} &= \ExP{}{\Tilde{x}_{k+1} \Tilde{x}_{k+1}^\intercal} \nonumber \\
    &\approx \frac{1}{N-1}\sum^{N-1}_{n=0}(x_{k+1}^n-\bar{x}_{k+1})(x_{k+1}^n-\bar{x}_{k+1})^\intercal,
\end{align}
where $x_{k+1}^n = f(x_k^n, u_k, w_k^n)$ is the state of the $n^\text{th}$ trajectory sample following the control sequence $\vu$ at time $k+1$. 

\subsubsection{Augmented belief-space system}

Since the costs $\phi(x_K)$ and $q(x_k)$ in the objective function \eqref{eqn:Shield MPPI objective} are only dependent on the state at a single time step, it is difficult to integrate the safety condition \eqref{eqn:SafeCondition} directly, which includes two consecutive system states. 
To this end, we introduce, for each $k=1,\ldots, K$, 
the augmented state $z_k = (z_k^{(1)},z_k^{(2)}) = (\hat{z}_{k},\hat{z}_{k-1}) \in \Rb^{2 n_z}$ and the corresponding augmented belief-space system.
\begin{equation}  \label{eqn:augmentedsystem}
\hspace*{-3mm}
z_{k+1} =
\begin{bmatrix}
z^{(1)}_{k+1}\\[5pt]
z^{(2)}_{k+1}
\end{bmatrix} 
=
\begin{bmatrix}
f_z(z^{(1)}_{k}, u_{k}) \\[2pt] 
z^{(1)}_{k}
\end{bmatrix}
=
\begin{bmatrix}
\hat{z}_{k+1}\\
\hat{z}_{k}
\end{bmatrix}.
\end{equation}

\subsection{Safety-aware Objective Function}\label{sec:safety_aware_objective_function_using_DCBF}
We can apply the MPPI algorithm to the augmented system \eqref{eqn:augmentedsystem} with cost,
\begin{align}\label{eqn:modified_objective}
    \min_\vv J(\vv) = 
    \mathbb{E}\left[ \phi_a(z_K) + \sum^{K-1}_{k=0} \left(q_a(z_k)  + \frac{\lambda}{2}v_k^\intercal \Sigma^{-1}_\epsilon v_k \right) \right],
\end{align}
to yield a sequence of optimal controls. Note that the step running cost $q_a(\cdot)$ is dependent on two consecutive states of the belief-space system \eqref{eqn:belief_space_system}. Taking a linear class-$\kappa$ function, it follows from \eqref{eqn:SafeCondition} that,
\begin{align}\label{eqn:DCBF_constraint2}
    h(\hat{z}_{k}) - (1-\beta)h(\hat{z}_{k-1}) \geq 0,
\end{align}
where $\beta \in (0,1)$, and $h(\cdot)$ is designed following \eqref{eqn:DCBFchoice0} and \eqref{eqn:DCBFchoice}. The resulting safe condition violation cost is 
\begin{equation}\label{eqn:cbfpenalty}
     C_\text{safe}(z_k) = C\, \max\{-h(z^{(1)}_k) + (1-\beta) h(z^{(2)}_k), 0\}.
\end{equation}
Hence, the state-dependent costs are 
\begin{subequations}\label{eqn:modified_state_dependent_costs}
\begin{align}
    q_a(z_k) = q(z_k^{(1)}) + C_\text{safe}(z_k),\\
    \phi_a(z_K) = \phi(z_K^{(1)}) + C_\text{safe}(z_K).
\end{align}
\end{subequations}
For convenience, we set $z_0^{(2)} = \hat{z}_{-1} = \hat{z}_0$, such that $C_\text{safe}(z_0) = 0$. The optimal control sequence can be calculated by \eqref{eqn:MPPI_update_law}, \eqref{eqn:Weights}, and \eqref{eqn:TrajectoryCost}, using \eqref{eqn:modified_state_dependent_costs} as the state-dependent running costs. The modified objective function \eqref{eqn:modified_objective} integrates the chance constraint \eqref{eqn:chance_constraint} by converting it into a cost minimization problem, which minimizes \eqref{eqn:BSS-MPPIobjective} while penalizing any violations of the safety condition \eqref{eqn:SafeCondition} to fulfill \eqref{eqn:chance_constraint}.

\section{Algorithm}

Algorithm~\ref{Algo:BSSMPPI} gives the pseudo-code for the proposed BSS-MPPI. Line~\ref{algo2:GetEstimateLine} obtaines the estimated robot state, lines~\ref{algo2:TrajectorySamplingBeginLine}-\ref{algo2:MPPITrajectoriesRolloutBeginLine} initialize states and trajectory costs, line \ref{algo2:ControlNoiseSampleLine}-\ref{algo2:WarmControl} sample control sequences, and lines~\ref{algo2:SystemPropagationLine}-\ref{algo2:CovariancePropagationLine} propagate the mean states and corresponding covariances. 

In \cite{CSVehiclePathPlanning, CCLTV, LCUGP}, the mean and covariance dynamics are modeled analytically and can be used to directly compute the next states. 
If the system is complex and highly nonlinear, such that the mean and covariance dynamics are difficult to model, lines~\ref{algo2:SystemPropagationLine}-\ref{algo2:CovariancePropagationLine} can be implemented by 
Algorithm~\ref{Algo:MeanAndCovariancePropagation}, which utilizes Monte-Carlo sampling to approximate mean and covariance propagation and is generally applicable. 

\begin{figure*}
    \centering
    \includegraphics[width=\textwidth]{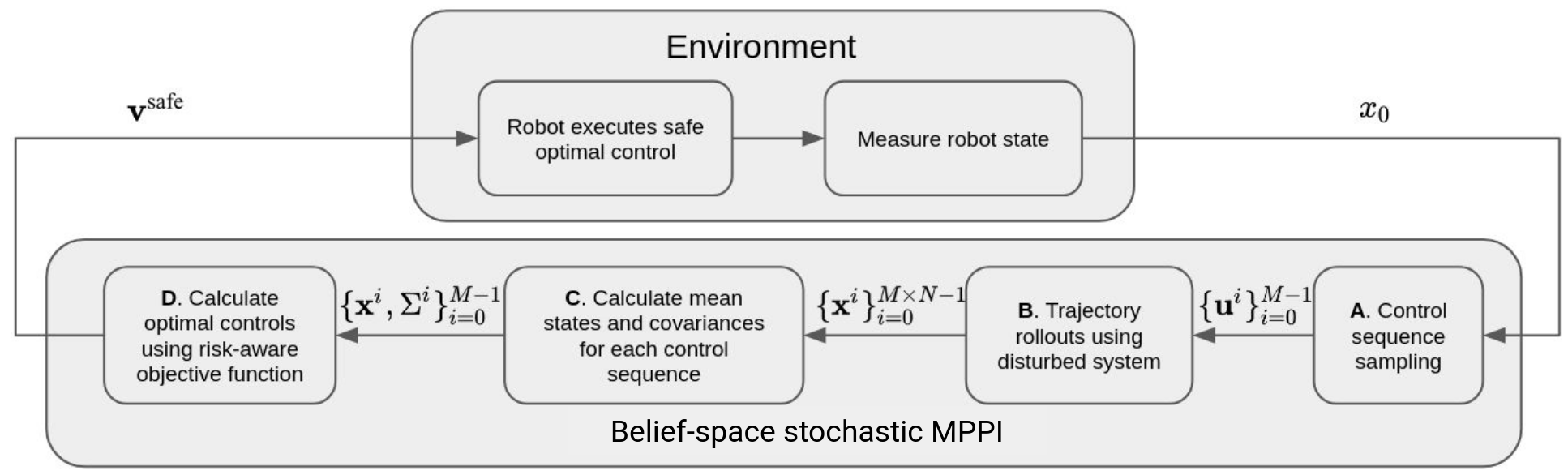}
    \caption{BSS-MPPI control architecture. }
    \label{fig:controlframework}
\end{figure*}

The disturbance $w_k$ in Algorithm~\ref{Algo:MeanAndCovariancePropagation} can be either from a fixed or a conditional (e.g., state-dependent) distribution. 
Lines~\ref{algo2:AugmentedBeliefSpaceState}-\ref{algo2:CalculateOptimalControlLine} in 
Algorithm~\ref{Algo:BSSMPPI} use the augmented belief space system \eqref{eqn:augmentedsystem} to evaluate the simulated trajectory costs $\Tilde{S}^m$, and compute the optimal controls using the control update law \eqref{eqn:MPPI_update_law}. 
Line~\ref{algo2:ExecuteCommandLine} sends the safe control command to the actuators, and line~\ref{algo2:InitializationLine} resets the nominal control sequence for the next control iteration. Fig.~\ref{fig:controlframework} shows the proposed architecture.

\begin{algorithm}
    \caption{Belief-space stochastic MPPI Algorithm}\label{Algo:BSSMPPI}
    \SetAlgoLined
    \LinesNumbered
    \SetKwInOut{Input}{Given}
    \Input{
    \noindent $\text{Shield-MPPI costs } q(\cdot), \phi(\cdot), \text{parameters} \; \gamma, \Sigma_\epsilon$;}
    \SetKwInOut{Input}{Input}
    \Input{
    \noindent $\text{Initial control sequence } \vv$}
    \While{task not complete}{
        $\bar{x}_{0}, \Sigma_{x_0} \leftarrow \textit{GetStateEstimate}()$;\\ \label{algo2:GetEstimateLine}
        \For{$m\leftarrow 0 \textbf{ to } M-1$ in parallel}{\label{algo2:TrajectorySamplingBeginLine} 
              $\bar{x}_0^m \leftarrow \bar{x}_0, \quad z^m_0 \gets [\bar{x}_0^\intercal, \text{vec}(\Sigma_{x_0}), \bar{x}_0^\intercal, \text{vec}(\Sigma_{x_0})]^\intercal, \quad \Tilde{S}^m \leftarrow 0$;\label{algo2:MPPITrajectoriesRolloutBeginLine}\\
              $\text{Sample }\bf{\epsilon}^m \leftarrow \{\epsilon_0^m,\ldots,\epsilon_{K-1}^m\}$;\label{algo2:ControlNoiseSampleLine}\\
              \For{$k\leftarrow 0 \textbf{ to } K-1$}{
                      $u_{k}^m \leftarrow v_{k} + \epsilon_k^m$;\label{algo2:WarmControl}\\
                      $\bar{x}_{k+1}^m \gets \bar{f}(\bar{x}_k^m , u_k^m)$; \label{algo2:SystemPropagationLine}\\
                      $\Sigma^m_{x_{k+1}} \gets f_{\Sigma}(\Sigma^m_{x_k}, u_k)$;\label{algo2:CovariancePropagationLine}\\
                      $z_{k+1}^m \gets [(\bar{x}_{k+1}^m)^\intercal, \text{vec}(\Sigma^m_{x_{k+1}}), (\bar{x}_k^m)^\intercal, \text{vec}(\Sigma^m_{x_{k}})]^\intercal$ \label{algo2:AugmentedBeliefSpaceState}\\
                      $\Tilde{S}^m \leftarrow \Tilde{S}^m + q(\bar{x}_k^m) + \gamma v_k^\intercal \Sigma_\epsilon^{-1}u_k^m +~C_\text{safe}(z_k^m)$;\label{algo2:TrajectoryCostOriginalLine}\\
                      }
                $\Tilde{S}^m \leftarrow \Tilde{S}^m + \phi(\bar{x}_K^m) + C_\text{safe}(z^m_K)$;\label{algo2:TrajectoryCostLine}\\ 
        } \label{algo2:TrajectorySamplingEndLine}
        $\vv^\text{safe} \gets \textit{OptimalControl}(\{\Tilde{S}^m\}_{m=0}^{M-1}, \{\vu^m\}_{m=0}^{M-1})$;\\\label{algo2:CalculateOptimalControlLine}
        % $\vv^\text{safe} \gets \textit{LocalRepair}(\bar{x}_0, \vv^+)$;\label{algo2:SafetyShieldLine}\\
        $\textit{ExecuteCommand}(v_0^\text{safe})$;\\ \label{algo2:ExecuteCommandLine} 
        $\vv \gets \vv^\text{safe}$;\label{algo2:InitializationLine}
    }
\end{algorithm}

\begin{algorithm}
    \caption{Mean and Covariance Propagation}\label{Algo:MeanAndCovariancePropagation}
    \SetAlgoLined
    \LinesNumbered
    \SetKwInOut{Input}{Given}
    \Input{
    \noindent $\text{System with disturbance } f(x, u, w)$, $\text{noise distribution } p(w|x).$}
    \SetKwInOut{Input}{Input}
    \Input{
    \noindent $\bar{x}_k, \Sigma_{x_k}, u_k$;}
    \For{$n\leftarrow 0 \textbf{ to } N-1$ in parallel}{\label{algo:TrajectorySamplingBeginLine} {
    \text{Sample $x^n_k \sim \mathcal{N}(\bar{x}_k, \Sigma_{x_k}), w_k \sim p(w_k|x_k)$}; \\
    $x^n_{k+1} = f(x^n_k, u_k, w_k)$\\
    }}
    $\bar{x}_{k+1} \gets \frac{1}{N}\sum^{N-1}_{n=0}x^n_{k+1}$\\
    $\Sigma_{x_{k+1}} \gets \frac{1}{N-1}\sum^{N-1}_{n=0}(x^n_{k+1}-\bar{x}_{k+1})(x_{k+1}^n-\bar{x}_{k+1})^\intercal$

\end{algorithm}

\section{Simulation Study}\label{Sec:Results}

In this section, we evaluate the proposed BSS-MPPI on a simulated vehicle-dynamics racing example, where the objective is to conclude a lap of a race course subject to minimizing a control objective. We develop the CBF-inspired heuristic for the considered application, and carry out a Monte-Carlo simulation study to evaluate the performance of the proposed BSS-MPPI (\textsc{bss-mppi}). We compare it with other state-of-the-art MPPI-based control approaches, in particular, the original MPPI (\textsc{mppi}, \cite{KLMPPI}) and the more recent S-MPPI (\textsc{s-mppi}, \cite{ShieldMPPI}).

\subsection{Experimental Setup}

\begin{figure}
    \centering
    \includegraphics{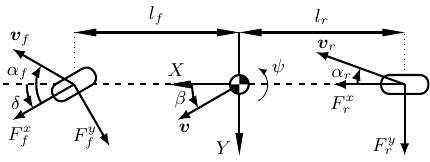}
    \vspace{-10pt}
    \caption{The planar single-track vehicle model used to model the AutoRally in Sec.~\ref{Sec:Results}.}
    \label{fig:SingleTrack}
    \vspace{-10pt}
\end{figure}

We use the AutoRally racing platform \cite{Autorally} to evaluate our method. 
The AutoRally is a 1m long, 0.4m wide electric vehicle with mass 22kg whose dynamics mimic a real vehicle. 
We model the dynamics using the discrete-time system  \eqref{DeterministicNonlinearSystem}, based on the planar single-track vehicle model \cite{Berntorp2014} sketched in Fig.~\ref{fig:SingleTrack}. The system state is $x=[v_X,v_Y,\dot \psi ,\omega_F,\omega_R, e_{\psi},e_{Y},s]^\top$, where $v_X$ is the longitudinal vehicle velocity, $v_Y$ is the lateral vehicle velocity, $\dot \psi$ is the yaw rate, $\omega_F,\omega_R$ is the front and rear wheel-speed, respectively, $e_Y$ is the lateral deviation from the centerline, $e_\psi$ is the yaw-angle deviation with respect to the centerline heading, and $s$ is the path coordinate in the road-aligned frame. The control input is $u=[\delta,T]^\top$ where $\delta$ is the steering angle at the front wheel and $T$ is the throttle. The tires are modeled by the Pacejka tire model and we use the friction ellipse to model combined slip \cite{Pacejka2006}.

We use the state-dependent running cost 
\begin{equation}\label{eq:CostFunction}
    q(x^m_k) = (x^m_k-x_g)^\top Q(x^m_k-x_g) + \boldsymbol{1}(x^m_k),
    \end{equation}
    where $Q=\mathrm{diag}(q_{v_X},q_{v_Y}, q_{\dot \psi},q_{\omega_F},q_{\omega_R}, q_{e_\psi},e_{y},q_s)$ is the cost-weight matrix, $x_g = \diag(v_g,0,\ldots,0)$ sets the target velocity, and 
    
    \begin{equation}
        \boldsymbol{1}(x^m_k) =  \left\{\begin{array}{ll}
            0, & \text{if } x^m_k \text{ satisfies the constraints},\\
            C_{\mathrm{obs}}, & \text{otherwise},
        \end{array}\right.
    \end{equation}
    is the collision penalty cost. Since \textsc{bss-mppi} extends \textsc{s-mppi} by introducing chance constraints into the problem formulation, our evaluation is mostly focused on the ability of  \textsc{bss-mppi} to satisfy the constraints under disturbed, that is, uncertain, dynamics. 
    We have executed 200 Monte-Carlo simulations where the vehicle is tasked to complete a lap of the AutoRally racetrack subject to zero-mean Gaussian process noise, $w_k \sim \mathcal{N}(0,\Sigma_w)$ with the reference velocity $v_g=6$~m/s. 
    
    To verify the performance sensitivity to the tuning of the objective function, we vary $q_{e_y}$ in \eqref{eq:CostFunction} while keeping the other variables fixed. 
    The reason is that the lateral deviation cost $q_{e_y}$ has a large impact on the maneuvering performance, and thereby it also has implications on the constraint satisfaction. 
    A relatively small $q_{e_y}$ favors cutting corners while a large $q_{e_y}$ makes the system stay close to the track centerline. Hence, a large $q_{e_y}$ reduces the risk of a collision against the track boundaries, but may lead to less efficient trajectories.

\subsection{Belief-Space Control Barrier Function Design}

Assuming that the racing track has constant width $2 w_{\rm T}$, we want to keep the vehicle's lateral deviation $e_y$ from the track centerline bounded by $|e_y| \leq w_{\rm T}$ for some given collision probability $\delta$.
Hence, we can write the chance constraints as
\begin{subequations}\label{eqn:track_chance_constraint1}
\begin{align}
 P(e_y < w_{\rm T}) > 1 - \epsilon,\\
 P(e_y > -w_{\rm T}) > 1 - \epsilon,
\end{align}
\end{subequations}
which keeps the chances of collision avoidance for each track boundary above $1-\epsilon$. 
 The chance constraints \eqref{eqn:track_chance_constraint1} are equivalent to
\begin{subequations}\label{eqn:track_chance_constraint2}
\begin{align}
 P(e_y \geq w_{\rm T}) \leq \epsilon, \label{Constraint1}\\
 P(e_y \leq -w_{\rm T}) \leq \epsilon. \label{Constraint2}
\end{align}
\end{subequations}
It follows from \eqref{eqn:chance_constraint2}, \eqref{eqn:deterministic_chance_constraint} that \eqref{eqn:track_chance_constraint2} can be converted to deterministic chance constraints,
\begin{subequations}
\begin{align}
\bar{e}_y - w_T + \nu  \sigma_y \leq 0, \label{eqn:deterministic_cc1}\\
\bar{e}_y + w_T -  \nu\sigma_y \geq 0, \label{eqn:deterministic_cc2}
\end{align}
\end{subequations}
where $\sigma_y$ is the standard deviation of $e_y$, obtained from the Monte-Carlo covariance propagation (see Algorithm~\ref{Algo:MeanAndCovariancePropagation}).
Since \eqref{eqn:deterministic_cc1} and \eqref{eqn:deterministic_cc2} are symmetric, we can combine them to form a single inequality,
\begin{align}
|\bar{e}_y| \leq w_T - \nu \sigma_y.    
\end{align}
Hence, we can formulate the  DCBF
\begin{align}\label{eqn:DCBF}
    h(x) = \left(w_T - \nu  \sigma_y\right)^2 - e_y^2,
\end{align}
and the corresponding safe set is \eqref{eqn:safeset}. We implement the chance constraints with the back-off coefficient $\nu = \sqrt{2}\mathrm{erf}^{-1}(1-2\epsilon)$ (see Remark~\ref{rem:1}), which for the considered application turned out to be a good compromise between conservativeness and safety.

\subsection{Simulation Results}

\begin{figure}
    \centering
    \includegraphics[scale=0.22]{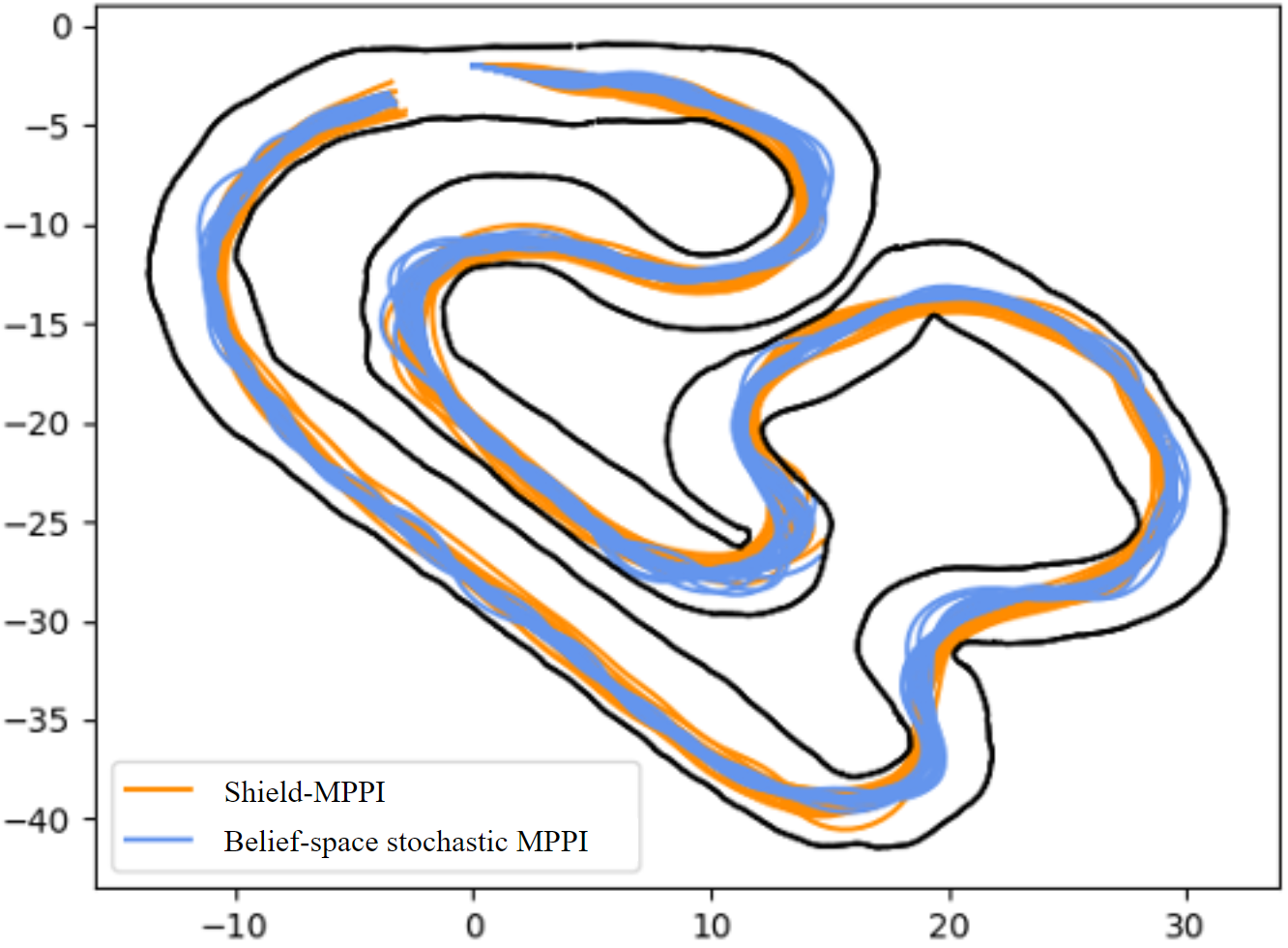}
    \caption{\textsc{s-mppi} and \textsc{bss-mppi} trajectory visualization for a set of  Monte-Carlo runs where the objective is to conclude a lap clock-wise. The trajectories generated by \textsc{bss-mppi} consider disturbances and are more conservative, resulting in lower speeds but fewer collisions}\label{fig:PathVisual}
\end{figure}

To illustrate, Fig.~\ref{fig:PathVisual} shows  the trajectories for a set of runs produced by \textsc{s-mppi} and \textsc{bss-mppi} when trying to manuever a lap of the race track. The trajectories produced by \textsc{s-mppi} have a tendency to go close to the track boundaries and on several occasions also cross the track boundaries, indicating  a subsequent crash. In contrast, owing to the chance constraints, \textsc{bss-mppi} trajectories tend to push closer to the track center, thus   yielding safer trajectories.

To quantify the results, we have executed  the different controllers using different tuning parameters, and for each set of parameters, we have executed the scenario in Fig.~\ref{fig:PathVisual} for 100 Monte-Carlo runs using different noise realizations and initial conditions.    Table~\ref{tab:1} shows the number of crashes and collisions, and the corresponding  ratios for different number of trajectories for \textsc{mppi}, \textsc{s-mppi}, and \textsc{bss-mppi}, respectively. We define a collision to occur when the vehicle deviates from the middle of the lane with more than 1.8m (which defines the chance constraint for the lateral deviation), and similarly we define a crash to occur when the vehicle goes outside of the (assumed constant) lanewidth $w_{\rm T}=2$m from the center of the lane, indicating a severe constraint violation. With this notion, we can quantify the number of constraint violations, in addition to quantifying the number of times the controllers heavily violates the constraints. 

Irrespective of the number of trajectories used, \textsc{bss-mppi} experiences substantially fewer crashes and collisions throughout the 100 Monte-Carlo runs. The vanilla MPPI (\textsc{mppi}) crashes at almost all  of the 100 Monte-Carlo runs and has on average 2.3 constraint violations per lap. The number of crashes and collisions for \textsc{s-mppi} is more or less constant irrespective of the number of trajectories, while the safety of \textsc{bss-mppi} improves as the number of trajectories involved to approximately solve the optimal control problem increases. With a sufficiently large $MN$, the collisions and crashes almost diminish. Even if the number of trajectories used to approximate the covariance propagation $N$ is large, \textsc{bss-mppi} is not able to recover appropriately if the number of sampled control sequences $M$ is small. 
%With a large $M$, it seems sufficient to add relatively few covariance and mean evaluation trajectories $N$. 

\begin{figure}
    \centering
    % {\includegraphics[width=\columnwidth,clip=true,trim={5 0 35 25}]{./figs/CrashRatePlot}}
    \includegraphics[width=\columnwidth,clip=true,trim={2 0 40 25}]{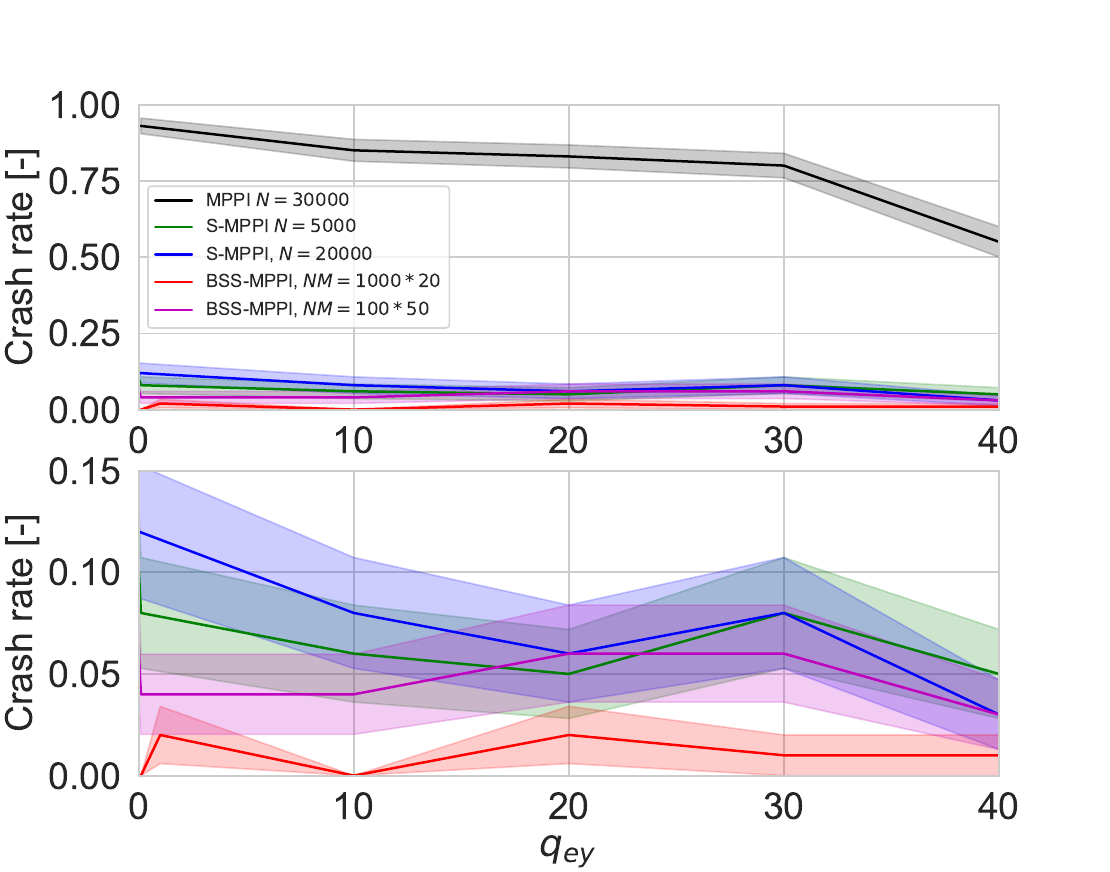}
    \vspace{-20pt}
    \caption{Crash-rate comparison between \textsc{bss-mppi} and \textsc{s-mppi} as function of $q_{e_y}$, with a zoom-in in the lower plot. The curves represent average performance over 100 Monte-Carlo runs with the shaded tubes showing the 1-$\sigma$ confidence intervals.}\label{fig:CrashRatePlot}
    \vspace{-20pt}
\end{figure}

\begin{figure}
    \centering
    \includegraphics[width=\columnwidth,clip=true,trim={10 0 40 15}]{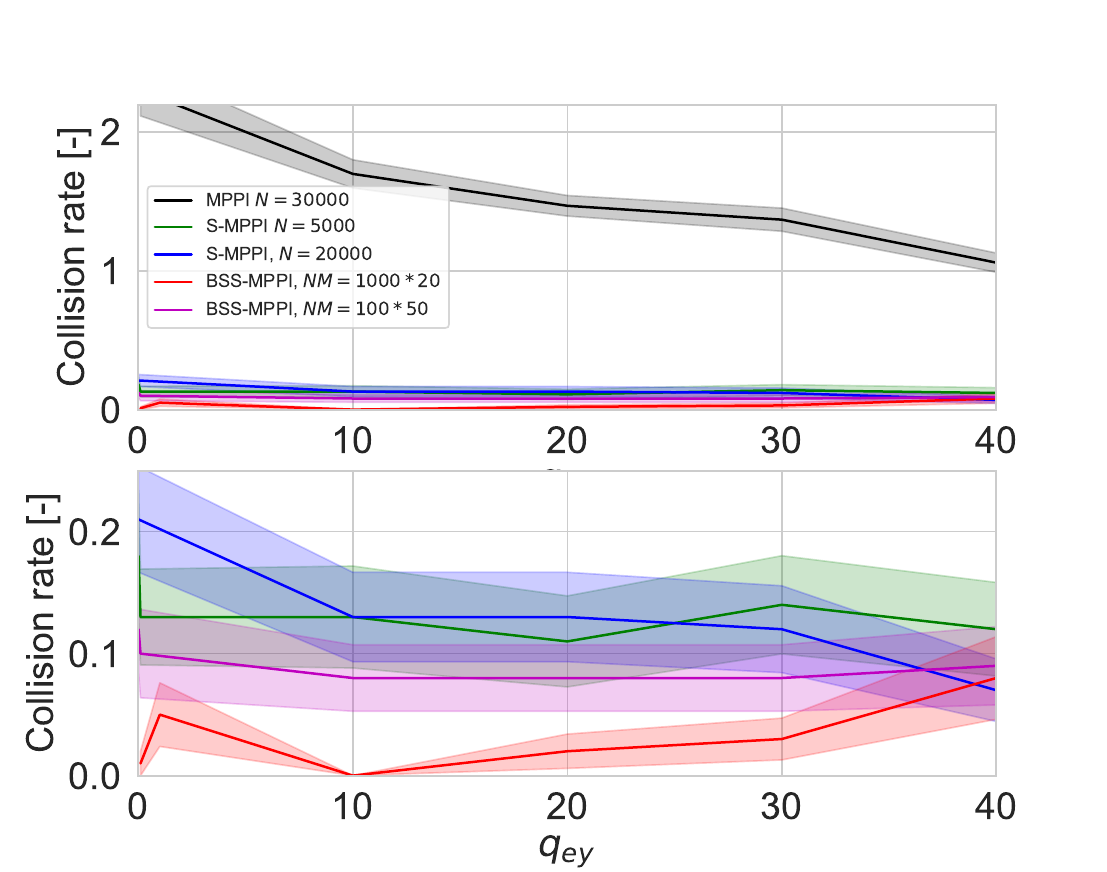}
    \vspace{-20pt}
    \caption{Collision-rate comparison between \textsc{bss-mppi} and \textsc{s-mppi} as function of $q_{e_y}$ corresponding to Fig.~\ref{fig:CrashRatePlot}. Since there can be multiple collisions (i.e., constraint violations) during a lap, the ratio can be greater than one.}\label{fig:CollisionhRatePlot}
    \vspace{-20pt}
\end{figure}

\begin{table} %
	\centering
	\caption{Crash ratio, collision ratio, and computational speed results for 100 Monte-Carlo runs  for a particular choice of cost function and $q_{e_y}=0.1$, and a control horizon of 20 steps. When a crash occurs, the particular Monte-Carlo run is terminated. The simulations use an Nvidia RTX2060 GPU and the computational speed is the average time it takes to execute the respective method one time step.}\label{tab:1}
	\vspace*{-5pt}
	\begin{tabular}{lccc}
		\toprule %
		Method &    Crashes & Collisions &  Speed [Hz]\\
		\midrule
        \textsc{mppi}, $M=30,000$  & 93\% & 230\% & 58.1\\
		\textsc{s-mppi}, $M=5,000$  & 8\% & 13\%&55.3\\
        \textsc{s-mppi}, $M=20,000$  & 8\% &  13\% &  42.9\\
        \textsc{bss-mppi}, $MN=100\cdot50$ & 4\% & 10\% &  43.1\\
        \textsc{bss-mppi}, $MN=500\cdot40$  & 0\% & 1\% &  39.7\\
        \textsc{bss-mppi}, $MN=1,000\cdot20$  & 0\% &  2\% & 39.1\\
			\bottomrule
	\end{tabular}
	\vspace*{-12pt}
\end{table}
To illustrate the sensitivity to cost function designs, Figs.~\ref{fig:CrashRatePlot}~and~\ref{fig:CollisionhRatePlot} display the crash and collision rates for different values of $M$ and $N$ as a function of $q_{e_y}$ (the lateral deviation cost term) along with the 1-$\sigma$ confidence interval. It is clear that  \textsc{bss-mppi}, the crash rate is relatively  insensitive to the value of $q_{e_y}$, while for  \textsc{s-mppi} with both $M=5000$ and $M=20,000$, the crash and collision rates generally reduce with increasing $q_{e_y}$. This is because s $q_{e_y}$ increases, the lateral deviation term dominates the cost function and for $q_{e_y}=40$, the controller tries to stay in the middle of the track and does not allow for closely tracking other objectives of the cost function. Furthermore, using $MN=5,000$ in \textsc{bss-mppi} in general leads to fewer crashes and collisions than \textsc{s-mppi} except for a very large penalization on the lateral deviation. However, again, such large penalization leads to other objectives of the cost function to  be ignored.
We can also see that the difference in crash rates between using $M=5,000$ and $M=20,000$ in \textsc{s-mppi} is small. Since  \textsc{s-mppi} does not account for any uncertainty, increasing the number of sampled control sequences $M$ will only help marginally in improving performance.

\begin{figure}
    \centering
    % {\includegraphics[width=\columnwidth,clip=true,trim={30 5 65 20}]{./figs/HeatMapBssMppi}}
    \includegraphics[width=\columnwidth,clip=true,trim={15 0 60 15}]{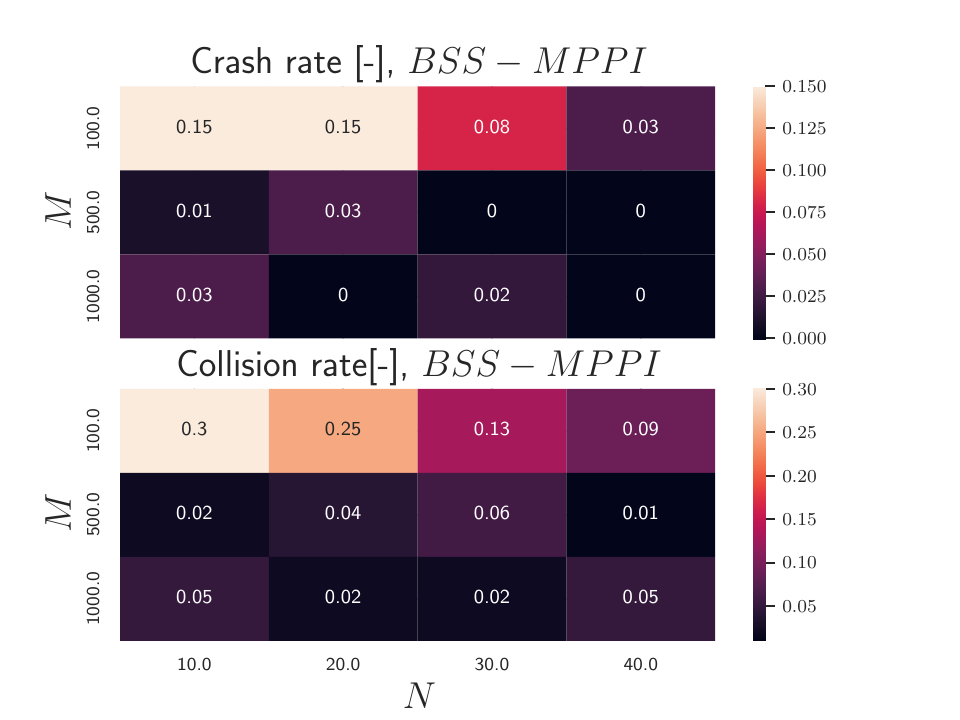}
    \vspace{-20pt}
    \caption{Crash and collision rates  as a function of the number of control sequences $M$ and covariance-propagation trajectories $N$, with $q_{e_y}=0.1$.}\label{fig:CrashRateHeatMap}
    \vspace{-20pt}
\end{figure}

Fig.~\ref{fig:CrashRateHeatMap} displays heatmaps of the crash and collision rates  to traverse a lap of the track using \textsc{bss-mppi} with different number of sampled control sequences $M$ and uncertainty  evaluation trajectories $N$. It is clear that increasing $N$ leads to fewer crashes, as the belief space trajectories and chance constraints are better approximated. However, only increasing $N$ is not sufficient, since using a small $M$ implies limited exploration of the optimal controls. Hence, the optimal choice is a trade-off between control exploration, uncertainty propagation, and computational resources.

%From the lower subplot in Fig.~\ref{fig:CrashRateHeatMap}, we note that increasing the number of control sequences $M$ improves velocity tracking. This is not surprising, since increasing $M$ implies enhancing the approximation quality of the optimal control problem, whereas the number $N$ is mostly related to constraint satisaction. 
\section{Conclusion} 
\label{sec:conclusion}

We presented a novel MPPI controller that accounts for system uncertainty by leveraging a CBF-inspired heuristic to satisfy chance constraints. The method can handle nonlinear dynamics and solves the underlying nonlinear stochastic optimal control problem by forward simulation of control and uncertainty trajectories, thus avoiding explicit linearization and optimization steps. The results as verified on a high-fidelity simulation environment indicate that our method effectively reduces the number of constraint violations, while at the same time achieving computational times comparable to previous MPPI approaches. 

In the future, we plan to evaluate the proposed method on an experimental setup, as well as extend it to perception-aware control.

\bibliographystyle{IEEEtran}
% \bibliography{references}
\bibliography{bib/references}

\end{document}